\documentclass{ws-jmrr}
\usepackage[sort,compress,super]{cite}

\usepackage{graphics, graphicx} 
\usepackage{amssymb}  
\usepackage{booktabs, multirow} 
\usepackage{soul}
\usepackage[table]{xcolor} 
\usepackage{changepage,threeparttable} 
\usepackage{amsmath}
\usepackage{algpseudocode}
\usepackage{algorithm}
\usepackage{adjustbox}
\usepackage{float}
\usepackage{microtype}
\usepackage{siunitx}
\usepackage{wrapfig}

\newcommand{\newstuff}[1]{\color{black}{#1}\color{black}}

\begin{document}

\catchline{0}{0}{2013}{}{}

\markboth{Shuyuan Yang, My H. Le, Kyle R. Golobish, Juan C. Beaver, Zonghe Chua}{Vision-Based Force Estimation for Minimally Invasive Telesurgery Through Contact Detection and Local Stiffness Models}

\title{Vision-Based Force Estimation for Minimally Invasive Telesurgery Through Contact Detection and Local Stiffness Models}

\author{Shuyuan Yang$^{a}$, My H. Le$^{a,b}$, Kyle R. Golobish$^c$, Juan C. Beaver$^b$, Zonghe Chua$^b$}

\address{$^a$Department of Computer and Data Sciences, Case Western Reserve University\\
10900 Euclid Ave Glennan Building Room 321, Cleveland, OH 44106, USA\\
E-mail: sxy841@case.edu}

\address{$^b$Department of Electrical, Systems and Computer Engineering, Case Western Reserve University
\\ 10900 Euclid Ave Glennan Building Room 321,
Cleveland, OH 44106, USA}

\address{$^c$Department of Mechanical and Aerospace Engineering, Case Western Reserve University
\\ 10900 Euclid Ave Glennan Building Room 479,
Cleveland, OH 44106, USA}

\maketitle

\begin{abstract}
In minimally invasive telesurgery, obtaining accurate force information is difficult due to the complexities of in-vivo end effector force sensing. This constrains development and implementation of haptic feedback and force-based automated performance metrics, respectively. Vision-based force sensing approaches using deep learning are a promising alternative to intrinsic end effector force sensing. However, they have limited ability to generalize to novel scenarios, and require learning on high-quality force sensor training data that can be difficult to obtain. To address these challenges, this paper presents a novel vision-based contact-conditional approach for force estimation in telesurgical environments. Our method leverages supervised learning with human labels and end effector position data to train deep neural networks. Predictions from these trained models are optionally combined with robot joint torque information to estimate forces indirectly from visual data. We benchmark our method against ground truth force sensor data and demonstrate generality by fine-tuning to novel surgical scenarios in a data-efficient manner. \newstuff{Our methods demonstrated greater than 90\% accuracy on contact detection and less than 10\% force prediction error. These results suggest potential usefulness of contact-conditional force estimation for sensory substitution haptic feedback and tissue handling skill evaluation in clinical settings.}


\end{abstract}


\begin{multicols}{2}
\section{INTRODUCTION}

\newstuff{In surgery, the ability to control applied tool-tissue forces is an essential skill for safe tissue handling. In minimally invasive telesurgery, haptic feedback has proven beneficial in this regard \cite{Wagner2007,Talasaz2017,patelHapticFeedbackForcebased2022}. However difficulties in implementing biocompatible, sterilizable, and miniaturized end-effector force sensing, have resulted in many systems lacking haptic feedback \cite{hadihosseinabadiForceSensingRobotassisted2022,patelHapticFeedbackForcebased2022}. Thus surgeons must undergo extensive training \cite{Herrell2005,Zorn2007,Abbas2018} and time-consuming evaluations \cite{martin1997osats,gohGlobalEvaluativeAssessment2012} to develop this crucial skill. While there have been efforts to automate skill evaluation, the difficulty of measuring force has resulted in no known automated tissue handling skill metrics \cite{chenObjectiveAssessmentRobotic2019}. 

The challenge of force estimation in minimally invasive telesurgery has motivated researchers to investigate indirect methods of estimating force. These include methods that use robot state \cite{yilmazNeuralNetworkBased2020}, visual information \cite{aviles2016towards,haouchineVisionbasedForceFeedback2018,jungVisionbasedSutureTensile2020,chuaForceEstimationRobotassisted2021}, or a combination of the two \cite{marbanRecurrentConvolutionalNeural2019,leeLearningEstimatePalpation2023}. Vision-based methods have shown promise, but still face limitations when adapting to new environments. Finite-element reconstruction methods \cite{haouchineVisionbasedForceFeedback2018} require knowledge of tissue attachment points and significant pre-processing of the stereoscopic video stream, while deep learning-based methods generalize poorly to visually dissimilar environments \cite{chuaForceEstimationRobotassisted2021}. Furthermore, these methods typically employ a supervised learning framework, and require high-quality ground truth force sensor data for training \cite{aviles2016towards,marbanRecurrentConvolutionalNeural2019,jungVisionbasedSutureTensile2020,chuaForceEstimationRobotassisted2021,leeLearningEstimatePalpation2023}. Such data is difficult to obtain in clinical settings, constraining researchers' abilities to collect enough data to train useful models for clinical deployment.

An increasingly accepted and scalable approach to overcoming the small data constraints in medical settings has been crowd-sourcing. This approach trades off the need for costly and precise measurement setups or scarce technical expertise of experts, for noisier but more voluminous data from non-experts. This has been done in domains such as pathology \cite{deng2023democrat}, as well as in surgical skill evaluation \cite{lendvayCrowdsourcingAssessSurgical2015,ohCrowdsourcedExpertEvaluations2018}. 

Here we present a versatile hybrid model- and learning-based approach to indirect force estimation that overcomes the challenge of collecting ground truth clinical force data for supervised learning. Inspired by ideas in crowd-sourced surgical skill evaluation, we leverage noisy but frequent measurements from non-expert human labelers. This is combined with imprecise robot sensor data, to estimate localized tissue contact, stiffness, and displacement. A contact-conditional local stiffness model is then used to provide an estimate of force based on displacement measurements. Adaptation to a new dataset can then be easily achieved using crowd-sourced human labels alone, with the added option of additional refinement if there is further access to robot sensor data. We extend the method to the common clinical situation in which intellectual property protections prevent accessing the robot state information in clinical settings. Under this constraint, we use only visual data (e.g. clinical video) to provide a normalized estimate of applied force. This latter output has exciting potential for quantifying tissue handling skill \cite{chenObjectiveAssessmentRobotic2019}, and can be scaled to provided sensory substitution force feedback \cite{kitagawaEffectSensorySubstitution2005,Talasaz2017,machacaROSbasedModularMultiModality2022,saracPerceptionMechanicalProperties2022}.}

\section{METHODS}

\subsection{Contact-conditional Local Force and Stiffness Estimation with Known Robot State Information}

We consider the case where the robot state is accessible, such as in a research robot like the da Vinci Research Kit\cite{kazanzides2014dvrk} (dVRK). A vision-based contact signal can be used with the robot end effector force $F_\text{PSM}\in\mathbb{R}^3$, and position measurements $p\in\mathbb{R}^3$, to derive an estimate of the effective stiffness $k$, of the material with which the end effector is in contact. The stiffness in the Z direction requires separate values to be fit for tension and compression. While in contact, we assume that at time $t$,
\begin{equation}
\label{eqn:stiffness_eqn}
F_{\text{PSM},t} \approx k^{(i)} s_t + c^{(i)} \text{\,,}
\end{equation}
where $F_\text{PSM}$ is the end effector estimated force \newstuff{in newtons } based on joint torque readings, $s_t = p_t-p_\tau$ is the end effector displacement \newstuff{in meters as measured } from the most recent onset of contact at time $\tau$, and $i$ is the $i^{\text{th}}$ demonstration. Both $k\in\mathbb{R}^3$ and $c\in\mathbb{R}^3$ were estimated for each demonstration using linear least squares \newstuff{with units of newton per meter and newtons, respectively}. Using the computed $k$, we then estimate the contact-conditional force at time $t$ for the $i^\text{th}$ demonstration as
\begin{equation}
\label{eqn:force_est}
F_{\text{computed},t} =
\begin{cases}
    k^{(i)} s_t, & \text{if in contact}\\
    0, & \text{otherwise}
\end{cases} \text{.}
\end{equation}
This method is henceforth called \textproc{C$_{\textproc{V}}$--K$_{\textproc{PSM}}$}. \newstuff{To benchmark our approach, we construct a best-case contact-conditional force estimate \textproc{C$_{\textproc{FS}}$--K$_{\textproc{FS}}$}. This uses the ground truth contact signal and the ground truth force to derive an estimate of $k$ and force. } To compare the contribution of the error from estimating $k^{(i)}$ from the noisy $F_{\text{PSM}}$ (as opposed to ground truth force), we also implemented an intermediate approach, \textproc{C$_{\textproc{V}}$--K$_{\textproc{FS}}$}. Here contact is estimated from vision, while $k^{(i)}$ is estimated from the ground truth force. Additionally, we compare against the classic position difference method, \textproc{PosDiff}, in which
\begin{equation}
\label{eqn:posdiff_eqn}
F_{\text{computed},t} = d^{(i)} (p_{\text{des},t}-p_t) + e^{(i)} \text{\,,}
\end{equation}
where $p_{\text{des},t}$ is the desired position of the end effector at time $t$ as reported by the dVRK. The scaling constant $d^{(i)}$ and offset $e^{(i)}$ for the $i^\text{th}$ demonstration are estimated through linear least squares with respect to $F_\text{PSM}$ using a similar assumption to Eqn.\,\ref{eqn:stiffness_eqn}.

\subsection{Contact-conditional Local Force Estimation with No Robot State Information}

\newstuff{When working with clinical versions of a telesurgical robot, the robot state information is often inaccessible due to intellectual property protections. Thus, surgical skills analysis and sensory substitution haptic augmentations in clinical settings often must rely purely on visual data streams \cite{5484176, giannarou2012probabilistic, ye2017self}. Here we accommodate this constraint in our force estimation approach. The measured stiffness constant is eliminated and a scaled measure of end effector position through vision is estimated in a viewpoint generalizable manner. This is similar to an existing approach that used estimated surgical tool path lengths for skill evaluation \cite{jin2018tool}. Even though the true force magnitude is not estimated, the scaled force variation can still provide a measure of tissue handling skill, and communicate performance-enhancing information through both force feedback \cite{quekSensorySub2015} and sensory substitution \cite{machacaROSbasedModularMultiModality2022,saracPerceptionMechanicalProperties2022} paradigms.}


We make the assumption that geometric and optical parameters do not vary substantially for standard telesurgical stereo endoscope and for particular types of surgical tools (i.e. EndoWrist Large Needle Drivers for a da Vinci surgical robot have the same geometries). An estimate of force can be achieved by first training a vision-based position estimator model in a supervised manner on a robot with access to state information. Alternatively, the robot can be instrumented with position measurement apparatus like infrared marker tracking. Once this initial training is done, the position estimator can be deployed on unseen systems, with the option of further benchtop fine-tuning. \newstuff{The position estimator is designed to learn and consequently generalize from data across varying viewpoints. To achieve this we normalized the position labels by the range of their corresponding demonstration example, such that}

\begin{equation}
    \label{eqn:normalized_pos}
    \newstuff{\hat{p}_t^{(i)} = \frac{p_t^{(i)}}{p_{\text{max}}^{(i)}-p_{\text{min}}^{(i)}} \text{\,,}}
\end{equation}

\parfillskip=0pt
\begin{figurehere}
\begin{center}
\centerline{\includegraphics[width=0.8\linewidth]{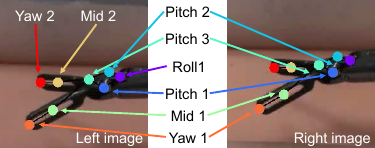}}
\caption{DeepLabCut labeled keypoints overlayed on both left and right camera views of the surgical manipulator.}
\label{figure:dlc-images}
\end{center}
\vspace{-2em}
\end{figurehere}

\noindent\newstuff{where $\hat{p}^{(i)}_t$ is the normalized position estimate of position $p^{(i)}_t$ at time $t$ for demonstration $i$. The variables $p^{(i)}_\text{max}$ and $p^{(i)}_\text{min}$ correspond to the maximum and minimum position attained in demonstration $i$.}

\newstuff{Training on this scaled position estimate results in a unitless position output from the position estimator. } These outputs $\hat{p}$, are then used instead of $p$ to compute $s_t$ in Eqn.\,\ref{eqn:force_est}, with $k^{(i)}$ being an arbitrary scaling constant. Thus the new equation is

\begin{equation}
    \label{eqn:normalized_pos_force}
    \newstuff{F_{\text{computed},t} = k^{(i)} (\hat{p_{t+1}}-\hat{p_t}) \text{\,.}}
\end{equation}


 This position estimator-based approach, henceforth called \textproc{FullVision}, does not require robot state information. For benchmarking, we use a priori knowledge of the ground truth force to fit the scaling constant using a similar assumption as in Eqn.\,\ref{eqn:stiffness_eqn}. This allowed for comparisons against the ground truth force measurement at similar scale.

\subsection{Vision-based Contact Detection}


To detect contact between the manipulator and tissue, we employed EfficientNetB3 \cite{tan2019efficientnet} as the feature encoder, coupled with a binary classification head. The model was trained using crowd-sourced contact labels which eliminate the need for force sensor data. To appropriately center a crop window of 234 by 234 pixels on the manipulator, we used our Normalized Position Estimator described in Section \ref{subsec:keypoints}. This centered the crop on the keypoint ``Mid 2" (Fig.\,\ref{figure:dlc-images}). During the training phase, including random rotations, cropping, flipping, erasing, and color jittering data augmentations were applied to the input images. 

 To validate our choice of a state-of-the-art network EfficientNetB3 model over a smaller network, we trained a small custom convolutional neural network model. This model consisted of 6 convolution layers with 8, 16, 32, 16,  8, and 4 channels, a kernel size of 3$\times$3, and stride 2 for the first layer, and stride 1 for all other layers. Average pooling layers with stride two were placed after every three convolution layers. A fully connected layer of 100 hidden units connected to a final binary classification layer was used. All activations were Rectified Linear Units (ReLU). We conducted a pseudo-randomized grid search to optimize the learning rate and L2 regularization weight. Both models were subjected to a training process spanning 150 epochs, with a batch size of 32, and were optimized using cross-entropy loss and the Adam optimizer. The model with the best performance on the validation set was chosen for evaluation.
 



\subsection{Vision-based Normalized Position Estimation}\label{subsec:keypoints}

\subsubsection{Keypoints Tracking}

In scenarios where access to robot kinematic and camera parameters data is not available, we devised an alternative approach to estimate a normalized 3D end effector position from video data based on extracted keypoints from DeepLabCut \cite{DLC}. We adopted the same keypoints as used in Lu et al. \cite{SuPer}, but introduced additional points at the middle of the jaws for a total of 8 keypoints for the tool. To fine-tune DeepLabCut for the pose estimation task, we randomly sampled 457 images from the training dataset. The output from DeepLabCut was a 32-dimensional vector corresponding to the pixel coordinates for each keypoint in a stereo image pair (Fig. \ref{figure:dlc-images}). Training was performed for 50,000 epochs on a Nvidia V100 graphics card. 

\subsubsection{Graph Neural Network Position Estimator}

The aim of our proposed method is to provide a generalizable and scalable approach to end effector position estimation that can be deployed off-the-shelf or fine-tuned quickly on a new robot. Thus, the resultant model must be data-efficient to train and fine-tune. To achieve this, we used a graph neural network (GNN) to model the fixed geometric relation of the detected keypoints as nodes on a graph. Directed edges between nodes were defined according to the end effector geometry. Next, eight undirected edges were added to connect corresponding nodes between stereo image pairs. The full graph architecture is shown in Fig.\,\ref{figure:nodes&edges}a. The input features vector of each node used a one-hot encoding of the static graph structure. This was concatenated with the horizontal and vertical normalized pixel coordinates to obtain an 18-dimensional vector. No temporal relationships between keypoints from consecutive images in a video stream were modeled. 


As shown in Fig.\,\ref{figure:nodes&edges}b, the GNN comprised two GraphSAGE \cite{GraphSage} convolution layers with 512 hidden units. A fully connected layer comprising 512 hidden units was placed between the GraphSAGE layers. All activation functions were ReLU. The model was trained to perform graph-level regression of three-dimensional position. Supervised learning was performed by collecting three-dimensional position labels for each stereo image training pair. Here we used the dVRK encoder-based end effector position
\parfillskip=0pt
\begin{figurehere}
\begin{center}
\centerline{\includegraphics[width=0.9\linewidth]{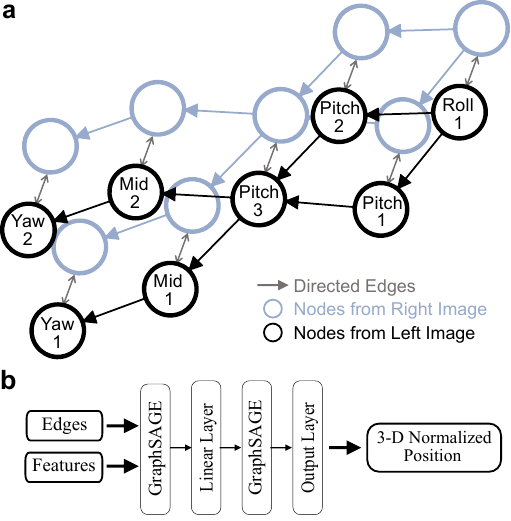}}
\caption{(a) Graph connectivity based on labeled keypoints from stereo images. (b) The inputs, layer arrangement, and final output of the graph neural network position estimator.}
\label{figure:nodes&edges}
\end{center}
\vspace{-2em}
\end{figurehere}
measurements, which were normalized by their range in each demonstration. We used a 4 by 4  hyperparameter grid search to obtain a learning rate and L2 regularization of $0.001$ and $0.0001$, respectively. Training was performed over 200 epochs using a batch size of 512. The chosen model was selected based on its performance on the validation set. 


\subsubsection{Fully Connected Neural Network Estimator}

To benchmark our GNN model, we used a custom Fully Connected Neural Network (FCN). This FCN had a symmetric architecture, comprising two identical sub-networks, one for each side of the stereo image pair. Each took as input the 2-dimensional pixel coordinates of the 8 keypoints identified through DeepLabCut as in Fig.\,\ref{figure:dlc-images}. This results in a 16-dimensional input vector. Each sub-network contained 4 fully connected layers with 16 hidden units with ReLU activation functions. The outputs from the sub-networks were fused using an additional fully connected layer with 32-dimensional input which then output a normalized 3-dimensional position. A hyperparameter grid search was performed to select the learning rate and L2 regularization, with the chosen values for both parameters being $0.0001$. Training was carried out for 200 epochs using a batch size of 32. As with the GNN, we selected the model with the best performance on the validation set.



\begin{figurehere}
\begin{center}
\centerline{\includegraphics[width=\linewidth]{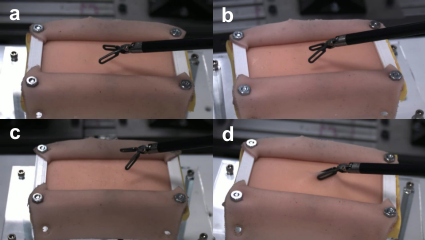}}
\caption{\newstuff{Sample images from the artificial silicone tissue dataset of the end effector in various states of contact for the four configurations with the largest camera offset.}}
\label{figure:silicone-dataset}
\end{center}
\vspace{-2em}
\end{figurehere}

\subsection{Datasets}

\subsubsection{Artificial Silicone Tissue Dataset}

We used a pre-existing dataset which consisted of 46 demonstrations of one dVRK patient-side manipulator (PSM) performing various retractions and palpation of manipulations on artificial silicone tissue. These were done under nine viewpoints and manipulator configurations (Fig.\,\ref{figure:silicone-dataset}). The dataset contained robot joint encoder current and desired positions, and current-based joint torque estimates. These data were collected at 1kHz. Stereo image pairs, each of size $960\times540$ were collected at 30Hz. The ground truth force data was collected from a 6-axis Nano17 sensor (ATI Automation, Apex, NC) placed underneath the tissue. The camera parameters were unknown, which is a typical constraint for \newstuff{clinical } data \cite{5484176}. This dataset was subsequently divided into training, validation and test sets. The training set and validation set comprised four configurations of 16 and 8 demonstrations containing a total of 56,098 and 28,107 examples, respectively. The test set contained demonstrations from the 4 training configurations and from 6 unseen configurations, resulting in a total of 22 demonstrations with 77,074 examples. 

We used Amazon Mechanical Turk to crowdsource visual contact training and validation labeled datasets on a downsampled and truncated version of the original dataset to reduce labeling costs. The downsampling factor was 45 and only the first 3000 images in each demonstration sequence were used for a total of 1,073 examples and 536 examples for the training and validation sets, respectively. Workers were shown examples of contact and no contact conditions and had to classify if the end effector was in contact with tissue. The final label for each example was averaged from the labels of five workers. This human-labeled dataset was used to train an ``MTurk" version of the vision-based contact detector.

To benchmark the quality of human labels, we generated contact labels from ground truth force sensor data by classifying force magnitudes of above 0.2\,N as being ``in contact". These sensor-labeled datasets was used to train a ground truth ``GT" version of the vision-based contact detector.

\subsubsection{Transfer Learning to Realistic Dataset}

To test the generality of our approach to a visually dissimilar dataset, we used a dataset comprising 40 demonstrations of either a left-side or right-side PSM being used on raw chicken skin wrapped around chicken thigh (See Fig.\,\ref{figure:realistic-dataset}). Scalability is achieved by only fine-tuning on a small subset of this new dataset using human-generated labels instead of sensor-based labels. Such fine-tuning approaches have previously proven effective in transferring surgical gesture classification from benchtop scenarios to clinical-like data \cite{itzkovichGeneralizationDeepLearning2022}. 12 demonstrations were used for training and the rest of the 28 demonstrations were used for testing. Both sets contain 21,073 and 49,374 images, respectively. To reduce labeling cost and training time, the dataset was downsampled by a factor of 45 and 10 for contact labeling and position estimation model training, respectively. This resulted in a total of 472 and 2,114 training examples, respectively.  The test set was not downsampled. Using these training sets, the vision-based contact estimators and position estimators that were previously trained on the silicone dataset were fine-tuned using the same hyperparameters and number of training epochs as before. Because hyperparameters were kept the same as those used with the silicone dataset, no validation set was required. The best model was selected based on its optimal training performance over all training epochs. To fine-tune the DeepLabCut keypoint detection, 90 images from the training set were sampled and consequently tuned over 20,000 epochs.

We conducted two experiments to test the generality of our proposed approach to novel surgical scenes. First, we benchmarked all contact detection, position estimation, and force estimation methods on the new dataset. Additionally for the position estimator, we separately tested the performance of the GNN and FCN when training from scratch on different amounts of data. This was done without pre-training on the silicone dataset. Second, we investigated the data efficiency of visual contact and position estimation when fine-tuning on new data. This was done by varying the amount of additional realistic data used during fine-tuning, and assessing model performance on the test set. 


\section{RESULTS AND DISCUSSION}

\subsection{Vision-based Contact Detection}


                           

The accuracy metrics for vision-based contact detection are shown in Table \ref{tab:contact_performance_silicone}. EfficientNet demonstrated consistently better performance regardless of the kind of training labels used. It achieved F1 scores of $0.985$ and $0.975$ when trained on force sensor-derived labels (GT), and human-derived labels (MTurk), respectively. In comparison, the small CNN achieved F1 scores of $0.979$ and $0.948$ on GT
\begin{figurehere}
\begin{center}
\centerline{\includegraphics[width=\linewidth]{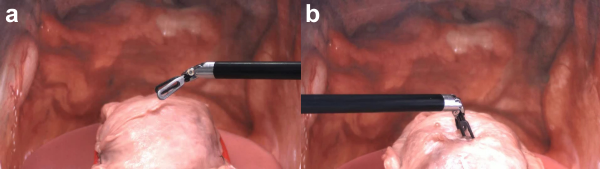}}
\caption{Sample images from the realistic dataset for (a) the right PSM configuration, and (b) the left PSM configuration.}
\label{figure:realistic-dataset}
\end{center}
\vspace{-2em}
\end{figurehere}
and MTurk labels, respectively. Comparing Fig.\,\ref{figure:contact-time-silicone}a and Fig.\,\ref{figure:contact-time-silicone}b, EfficientNet's performance is superior to the small CNN, matching the ground truth contact signal more consistently. As shown by the non-thresholded predictions (dotted lines in Fig.\,\ref{figure:contact-time-silicone}), the small CNN prediction confidence fluctuated frequently, resulting in misclassifications especially during periods when contact was just made or just broken. The agreement of the performance of the models trained on the ground truth labels, and the MTurk labels, indicates that the use of human labels is comparable to using an actual force sensor to detect contact. Furthermore, there it has the added advantage being usable with \newstuff{clinical } datasets that do not have accurate force measurements.

\subsection{Position Estimation Methods}

\newstuff{Table \,\ref{tab: nn_rmse} presents the accuracy metrics for the normalized position estimator. The error in the test set is reported in the normalized unitless scale. This represents a percentage error with respect to the distance traversed by the end-effector over the corresponding demonstration. For interpretability, Table \,\ref{tab: nn_rmse} also reports RMSE errors at the scale of the test set demonstrations.} 

The results in Table \,\ref{tab: nn_rmse} illustrate that both the normalized position estimators -- the GNN model and the FCN model -- exhibited comparable performance on the silicone dataset. The GNN model demonstrated an approximately $2\%$ lower accuracy compared to the FCN model across all axes of force. This reduction in accuracy is expected given the shallow network structure of the GNN. This constraint is imposed by the sparseness of the geometry-based graph structure used, where adding more GraphSAGE layers would result in redundant messages being passed between nodes.

The comparison of visual predictions and actual positions, as depicted in Fig.\,\ref{figure:nn-plot}, confirms that both models are able to model the movement trends of the ground truth end effector position. Both models more accurately captured the X and Z axes movements compared to the Y-axis movements. X and Z correspond to the left-right and up-down directions in the stereo images. Y corresponds to depth in the stereo image, which contains more ambiguity. \newstuff{When converted back into the scale of the test set, the accuracy of the estimates is within 5\,mm. Since the position estimates are used to generate force estimates, the acceptability of positional accuracy will be evaluated based on force prediction accuracy, which we describe in later sections.}

\begin{table*}
\tbl{Contact detection model performance metrics. \label{tab:contact_performance_silicone}}
{\begin{adjustbox}{max width=0.9\textwidth}
\begin{tabular}{@{}llccccccccc@{}}
\toprule
\multirow{2}{*}{\textbf{Model}} & \multirow{2}{*}{\textbf{Label}} & \multicolumn{4}{c}{\textbf{Silicone}} & & \multicolumn{4}{c}{\textbf{Realistic}} \\
\cline{3-6} \cline{8-11}
 & & \textbf{Accuracy} & \textbf{Precision} & \textbf{Recall} & \textbf{F1 Score} & & \textbf{Accuracy} & \textbf{Precision} & \textbf{Recall} & \textbf{F1 Score}\\ 
\colrule

\multirow{4}{*}{small CNN} & \multirow{2}{*}{GT}   & 0.964 & 0.982 & 0.975 & 0.979 & & 0.820 & 0.863 & 0.853 & 0.855 \\
                           &                       & ± 0.010 & ± 0.006 & ± 0.011 & ± 0.007 & & ± 0.050 & ± 0.049 & ± 0.086 & ± 0.050 \\ 
                           \cline{2-11}
                           & \multirow{2}{*}{MTurk} & 0.916 & 0.992 & 0.908 & 0.948 & & 0.848 & 0.861 & 0.909 & 0.882 \\
                           &                       & ± 0.022 & ± 0.003 & ± 0.025 & ± 0.014 & & ± 0.043 & ± 0.049 & ± 0.059 & ± 0.039 \\ 
\hline
                           
\multirow{4}{*}{EfficientNet} & \multirow{2}{*}{GT}  & 0.975 & 0.983 & 0.988 & 0.985 & & 0.875 & 0.844 & 0.988 & 0.909 \\
                              &                      & ± 0.005 & ± 0.007 & ± 0.008 & ± 0.003 & & ± 0.047 & ± 0.058 & ± 0.018 & ± 0.035 \\ 
                              \cline{2-11}
                              & \multirow{2}{*}{MTurk} & 0.959 & 0.990 & 0.960 & 0.975 & & 0.899 & 0.873 & 0.986 & 0.925 \\
                              &                      & ± 0.018 & ± 0.003 & ± 0.023 & ± 0.012 & & ± 0.041 & ± 0.053 & ± 0.010 & ± 0.031 \\ 
\botrule
\end{tabular}
\end{adjustbox}}
\end{table*}

\begin{table*}
\newstuff{
\tbl{Position error metrics for vision-based normalized position estimators. \label{tab: nn_rmse}}
{\begin{adjustbox}{max width=\linewidth}
\begin{tabular}{@{}llccccccccc@{}}
\toprule
\multirow{2}{*}{\textbf{Dataset}} & \multirow{2}{*}{\textbf{Model}} & \multicolumn{4}{c}{\textbf{RMSE -- Normalized Position (\%)}} & & \multicolumn{4}{c}{\textbf{RMSE -- Rescaled Position (m)*}}\\
\cline{3-6} \cline{8-11}
 &  & \textbf{Overall} & \textbf{$X$} & \textbf{$Y$} & \textbf{$Z$} &  & \textbf{Overall} & \textbf{$X$} & \textbf{$Y$} & \textbf{$Z$}\\
\colrule
\multirow{4}{*}{Silicone} & \multirow{2}{*}{GNN} & 0.087 & 0.052 & 0.121 & 0.089 & & 0.002 & 0.002 & 0.003 & 0.003 \\
 &  & ± 0.017 & ± 0.012 & ± 0.024 & ± 0.016 & & ± 0.001 & ± 0.001 & ± 0.001 & ± 0.001 \\
\cline{2-11}
 & \multirow{2}{*}{FCN} & 0.085 & 0.047 & 0.122 & 0.085 & & 0.002 & 0.001 & 0.003 & 0.003 \\
 &  & ± 0.020 & ± 0.011 & ± 0.031 & ± 0.017 & & ± 0.001 & ± 0.000 & ± 0.001 & ± 0.001 \\
\hline

\multirow{4}{*}{Realistic} & \multirow{2}{*}{GNN} & 0.091 & 0.041 & 0.141 & 0.093 & & 0.004 & 0.002 & 0.005 & 0.003 \\
&  & ± 0.027 & ± 0.014 & ± 0.038 & ± 0.030 & & ± 0.001 & ± 0.000 & ± 0.001 & ± 0.001 \\
\cline{2-11}
& \multirow{2}{*}{FCN} & 0.088 & 0.043 & 0.133 & 0.087 & & 0.003 & 0.002 & 0.005 & 0.003 \\
&  & ± 0.027 & ± 0.014 & ± 0.040 & ± 0.028 & & ± 0.001 & ± 0.000 & ± 0.001 & ± 0.001 \\
\botrule
\multicolumn{11}{l}{\small \newstuff{*Normalized Position RMSE rescaled into the range of the test set for interpretability.}} \\
\end{tabular}
\end{adjustbox}}
}
\end{table*}

\begin{figurehere}
\begin{center}
\centerline{\includegraphics[width=0.9\linewidth]{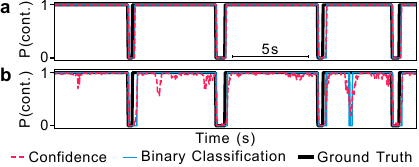}}
\caption{\newstuff{Predicted contact probabilities for the (a) EfficientNetB3 and (b) the small CNN model, on one demonstration from the silicone dataset. All models were trained on human contact labels.}}
\label{figure:contact-time-silicone}
\end{center}
\vspace{-3em}
\end{figurehere}

\begin{figurehere}
\begin{center}
\centerline{\includegraphics[width=0.8\linewidth]{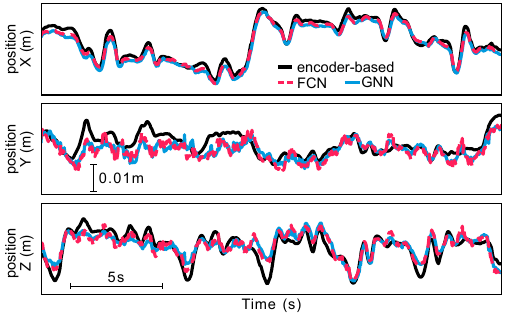}}
\caption{\newstuff{Normalized end-effector position predictions of the graph neural network (GNN) and fully connected network (FCN) compared to the joint encoder-based position.}}
\label{figure:nn-plot}
\end{center}
\vspace{-4em}
\end{figurehere}

\subsection{Contact-conditional Local Force and Stiffness Estimation with Known Robot State Information}

\subsubsection{Model-based Stiffness Estimation}

\newstuff{The average estimated stiffnesses of the manipulated materials are reported in Table \ref{tab: stiffness}. As it was derived from force sensor data, the estimated stiffness from \textproc{C$_{\textproc{FS}}$--K$_{\textproc{FS}}$} functions as the ground truth reference stiffness. Comparing this estimate against \textproc{C$_{\textproc{V}}$--K$_{\textproc{PSM}}$}, the difference in the mean stiffness were $-44$, $+37$, $+1$, $-10$\,$\text{Nm}^{-1}$ in the $X$, $Y$, $Z^+$, and $Z^-$ directions, respectively. Thus, the average error was 13\% across all directions, with a maximum error of 26\% in the X direction. This is comparable to the limits of human stiffness discrimination without visual feedback, which has a Weber Fraction of 23\% \cite{jonesPerceptual1990}. However, it is above 14\% Weber Fraction for stiffness discrimination with visual feedback \cite{varad2008}. This suggests that the contact conditional stiffness estimation approach } \newstuff{is promising, but does require a more accurate estimate of force to facilitate tissue differentiation tasks. } Fig.\,\ref{figure:R3_M1_T1_1_F_vs_x} shows a representative example of the fitted stiffness values based on the different sources of force data.

\subsubsection{Force Estimation}
\label{sec:force_estimation}

\newstuff{Table \ref{tab: rmse_silicone} presents the average normalized root mean square error (NRMSE) of the predicted force. This is computed} 
\begin{figurehere}
\begin{center}
\centerline{\includegraphics[width=\linewidth]{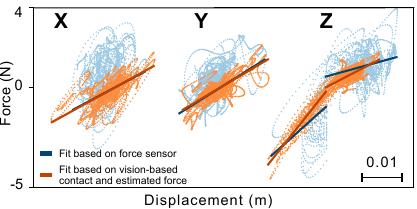}}
\caption{Best fit stiffness models based on either force sensor readings, or the estimated end effector forces using joint torques, with contact conditional displacement readings from joint encoders, for one representative example of the silicone dataset. Dots represent individual data points.}
\label{figure:R3_M1_T1_1_F_vs_x}
\end{center}
\vspace{-2em}
\end{figurehere}
\begin{tablehere}
\newstuff{
\tbl{Estimated stiffness of manipulated materials. \label{tab: stiffness}}
{\begin{adjustbox}{max width=\linewidth}
\begin{tabular}{@{}llcccc@{}}
\toprule
\multirow{2}{*}{\textbf{Dataset}} & \multirow{2}{*}{\textbf{Model}} & \multicolumn{4}{c}{\textbf{Stiffness ($\text{Nm}^{-1}$)}} \\
\cline{3-6}
 &  & \textbf{$X$} & \textbf{$Y$} & \textbf{$Z^+$} & \textbf{$Z^-$} \\
\colrule
\multirow{6}{*}{Silicone} & \multirow{2}{*}{\textproc{C$_{\textproc{FS}}$--K$_{\textproc{FS}}$}} & 168 & 182 & 108 & 332 \\
 &  & ± 47 & ± 44 & ± 25 & ± 52 \\
\cline{2-6}
 & \multirow{2}{*}{\textproc{C$_{\textproc{V}}$--K$_{\textproc{FS}}$}} & 170 & 185 & 107 & 335 \\
 &  & ± 46 & ± 44 & ± 25 & ± 51 \\
\cline{2-6}
 & \textproc{C$_{\textproc{V}}$--K$_{\textproc{PSM}}$} & 124 & 219 & 109 & 322 \\
 & (our approach) & ± 41 & ± 60 & ± 58 & ± 104 \\
\hline
\multirow{6}{*}{Realistic} & \multirow{2}{*}{\textproc{C$_{\textproc{FS}}$--K$_{\textproc{FS}}$}} & 126 & 131 & 96 & 245 \\
 &  & ± 36 & ± 55 & ± 29 & ± 148 \\
\cline{2-6}
 & \multirow{2}{*}{\textproc{C$_{\textproc{V}}$--K$_{\textproc{FS}}$}} & 126 & 131 & 98 & 246 \\
 &  & ± 36 & ± 55 & ± 33 & ± 148 \\
\cline{2-6}
 & \textproc{C$_{\textproc{V}}$--K$_{\textproc{PSM}}$} & 94 & 119 & 135 & 239 \\
 & (our approach) & ± 78 & ± 105 & ± 88 & ± 124 \\
\botrule
\end{tabular}
\end{adjustbox}}
}
\end{tablehere}
\newstuff{with respect to the ground truth force sensor measurements over all test demonstrations. The top rows present contact conditional methods that use robot position information, with different sources of contact and force information: \textproc{C$_{\textproc{FS}}$--K$_{\textproc{FS}}$}, \textproc{C$_{\textproc{V}}$--K$_{\textproc{FS}}$}, and \textproc{C$_{\textproc{V}}$--K$_{\textproc{PSM}}$}. These are benchmarked against $F_\text{PSM}$ and $\textproc{PosDiff}$ force estimates. The NRMSE in each force direction calculated element-wise as
\begin{equation}
\label{eqn:nrmse}
\text{NRMSE} =  \frac{ \sqrt{ \frac{1}{N} \sum^{N}_{n=1} (F_{\text{computed},n}^{(i)}-F_n^{(i)})^2 } }{F_\text{max}^{(i)} - F_\text{min}^{(i)}}                    \text{,}
\end{equation}
where $F_{\text{computed}}$ is the computed force estimate, $F$ is the ground truth force as measured from the force sensor, $F_\text{max}$ is the maximum force observed, $F_\text{min}$ the minimum force, and $N$ is the number of data points, in the $i^\text{th}$ demonstration. \cite{yilmazNeuralNetworkBased2020,Pique2019dynamic}}

\textproc{C$_{\textproc{V}}$--K$_{\textproc{PSM}}$} showed lower mean NRMSE in all directions compared to force estimates based on joint torques (F$_\textproc{PSM}$). \textproc{C$_{\textproc{V}}$--K$_{\textproc{PSM}}$} also outperforms \textproc{PosDiff} which is a traditional approach to providing a scaled form of haptic feedback. The advantage of \textproc{C$_{\textproc{V}}$--K$_{\textproc{PSM}}$} is that it is less sensitive to the internal manipulator dynamics that affect F$_\textproc{PSM}$ and \textproc{PosDiff}. \newstuff{Critically, the NRMSE of the norm (i.e. magnitude) and in each direction of \textproc{C$_{\textproc{V}}$--K$_{\textproc{PSM}}$} on the silicone dataset is below the 10\% scaling threshold identified by Huang et al. \cite{huang2020characterizing} for degraded teleoperated palpation. This threshold also corresponds to the average human force JND of 10\% \cite{Feyzabadi2013}. Our results thus indicate that contact-conditional force estimation for force feedback has potential to improve telesurgical manipulation.}

The increase in error between \textproc{C$_{\textproc{FS}}$--K$_{\textproc{FS}}$} and \textproc{C$_{\textproc{V}}$--K$_{\textproc{FS}}$} was smaller than that between \textproc{C$_{\textproc{V}}$--K$_{\textproc{FS}}$} and \textproc{C$_{\textproc{V}}$--K$_{\textproc{PSM}}$}. This suggests that there was a larger error contribution from the stiffness estimation (K$_{\textproc{FS}}$ vs. K$_{\textproc{PSM}}$) than from the contact detection (C$_{\textproc{FS}}$ vs. C$_{\textproc{V}}$).

\begin{figure*}
\begin{center}
\centerline{\includegraphics[width=\linewidth]{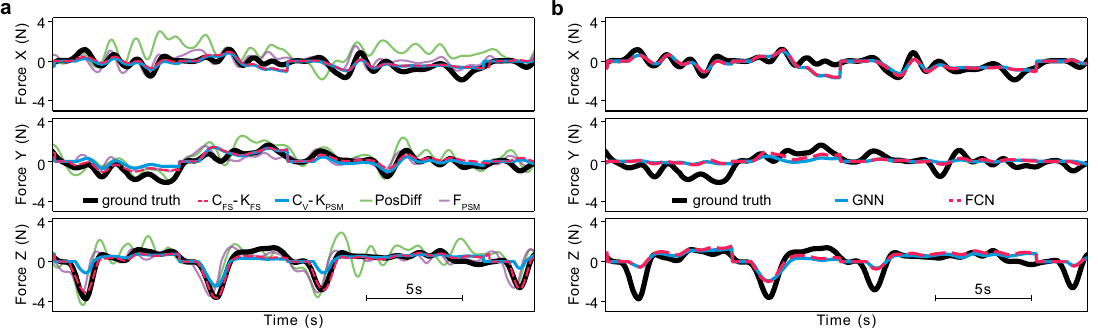}}
\caption{\newstuff{Example force predictions for force estimation approaches that require (a) robot state, and (b) no robot state information, for one demonstration from the silicone dataset.}}
\label{figure:R3_M1_T1_1}
\end{center}
\vspace{-2em}
\end{figure*}

The large increase in error from \textproc{C$_{\textproc{V}}$--K$_{\textproc{FS}}$} to \textproc{C$_{\textproc{V}}$--K$_{\textproc{PSM}}$} in the Z direction was likely due to the higher overall stiffness in the Z$^-$ direction. In Fig.\,\ref{figure:R3_M1_T1_1}, \textproc{C$_{\textproc{V}}$--K$_{\textproc{PSM}}$} shows general tracking of force variation. However it displays underestimation in the Z$^-$ direction at high compression forces compared to \textproc{C$_{\textproc{FS}}$--K$_{\textproc{FS}}$}.

\newstuff{The \textproc{C$_{\textproc{V}}$--K$_{\textproc{PSM}}$} force estimates in Fig.\,\ref{figure:R3_M1_T1_1} show occasional instances of poor contact classifications that caused the contact condition to change abruptly. This resulted in the predicted force decreasing to zero sharply instead of smoothly like with the ground truth. These abrupt force deviations can potentially cause tissue damage if teleoperating with direct force feedback. Strategies to eliminate this include implementing a smoothing filter on the force.
Given that direct force feedback also has to contend with the safety concerns of control instability \cite{Hashtrudi-Zaad2001}, we identify haptic sensory substitution \cite{kitagawaEffectSensorySubstitution2005,Talasaz2017,machacaROSbasedModularMultiModality2022,saracPerceptionMechanicalProperties2022} as the more promising use case for the contact-conditional visual force estimates.} 

\begin{table*}
\newstuff{
\tbl{Normalized RMSE of force estimates of different force estimation methods with respect to force sensor measurements. \label{tab: rmse_silicone}}
{\begin{adjustbox}{max width=\textwidth}
\begin{tabular}{@{}lccccccccc@{}}
\toprule
\multirow{3}{*}{\textbf{Method}} & \multicolumn{9}{c}{\textbf{NRMSE (\%)}} \\
\cline{2-10}
& \multicolumn{4}{c}{\textbf{Silicone}} & & \multicolumn{4}{c}{\textbf{Realistic}} \\
\cline{2-5} \cline{7-10}
& \textbf{Norm} & \textbf{$X$} & \textbf{$Y$} & \textbf{$Z$}  & & \textbf{Norm} & \textbf{$X$} & \textbf{$Y$} & \textbf{$Z$} \\
\colrule
\newstuff{\textit{with robot state}} \\
\hline
\multirow{2}{*}{\textproc{F}$_\textproc{PSM}$} & 0.079 & 0.192 & 0.141 & 0.129 & & 0.198 & 0.330 & 0.247 & 0.265 \\
& ± 0.017 & ± 0.032 & ± 0.043 & ± 0.034 & & ± 0.088 & ± 0.119 & ± 0.082 & ± 0.137 \\
\hline
\multirow{2}{*}{\textproc{C$_{\textproc{FS}}$--K$_{\textproc{FS}}$}} & 0.052 & 0.107 & 0.092 & 0.052 & & 0.060 & 0.072 & 0.097 & 0.080 \\
& ± 0.007 & ± 0.017 & ± 0.016 & ± 0.009 & & ± 0.011 & ± 0.014 & ± 0.018 & ± 0.025 \\
\hline
\multirow{2}{*}{\textproc{C$_{\textproc{V}}$--K$_{\textproc{FS}}$}} & 0.053 & 0.107 & 0.092 & 0.055 & & 0.060 & 0.072 & 0.097 & 0.081 \\
& ± 0.008 & ± 0.017 & ± 0.016 & ± 0.010 & & ± 0.011 & ± 0.014 & ± 0.018 & ± 0.025 \\
\hline
\textproc{C$_{\textproc{V}}$--K$_{\textproc{PSM}}$} & 0.068 & 0.106 & 0.090 & 0.090 & & 0.097 & 0.107 & 0.136 & 0.149 \\
(our approach) & ± 0.015 & ± 0.012 & ± 0.021 & ± 0.022 & & ± 0.042 & ± 0.030 & ± 0.039 & ± 0.107 \\
\hline
\multirow{2}{*}{\textproc{PosDiff}} & 0.088 & 0.127 & 0.118 & 0.129 & & 0.132 & 0.153 & 0.170 & 0.189 \\
& ± 0.007 & ± 0.016 & ± 0.023 & ± 0.016 & & ± 0.052 & ± 0.059 & ± 0.060 & ± 0.084 \\
\colrule
\newstuff{\textit{without robot state *}} \\
\hline
\multirow{2}{*}{\textproc{FullVision (GNN)}} & 0.077 & 0.115 & 0.126 & 0.093 & & 0.081 & 0.085 & 0.130 & 0.108 \\
& ± 0.013 & ± 0.014 & ± 0.015 & ± 0.022 & & ± 0.012 & ± 0.014 & ± 0.025 & ± 0.030 \\
\hline
\multirow{2}{*}{\textproc{FullVision (FCN)}} & 0.076 & 0.116 & 0.121 & 0.093 & & 0.081 & 0.085 & 0.127 & 0.103 \\
& ± 0.013 & ± 0.014 & ± 0.015 & ± 0.026 & & ± 0.010 & ± 0.015 & ± 0.021 & ± 0.024 \\
\botrule
\multicolumn{10}{l}{\small *Force estimates were rescaled to match force scaling of the test data to allow for interpretable comparison} \\
\end{tabular}
\end{adjustbox}}
}
\end{table*}

\subsection{Contact-conditional Local Force Estimation with No Robot State Information}

The last two rows of Table 4 present the NRMSE of the force estimation methods with no robot state information. Here, the unitless force estimates are rescaled to match that of the test set for interpretability. The accuracies of both the GNN and FCN vision-only force estimation methods are shown to be comparable to \newstuff{those } using F$_{\textproc{PSM}}$, a method that requires robot state information. In the Y direction, there is
notable force understimation. This error can be largely attributed to the low positional accuracy of the normalized position estimates in the Y direction (Table\,\ref{tab: nn_rmse} and Fig.\,\ref{figure:nn-plot}). Despite this, both GNN and FCN methods show tracking of force variations, indicating the viability of such methods for obtaining a general measure of tissue handling force. 

\newstuff{The rescaling used linear least squares to tune the stiffness parameter to best match ground truth. This came at the expense of presenting more force variation. Alternatively, we can improve presentation of force variation by increasing the stiffness parameter and trading off some accuracy. Like in Section \ref{sec:force_estimation}, haptic sensory substitution is a highly viable method of presenting such force feedback. When used in this manner, representation of the force is now arbitrarily scaled such that accurately tracking relative force variations is more important than estimating exact force magnitudes.}


\subsection{Generality of Approach to Novel Surgical Scenes}


                           
Table \ref{tab:contact_performance_silicone} presents the performance metrics for each fine-tuned contact detection model on the realistic dataset. When trained on MTurk labels, EfficientNet exhibited a decrease in F1 score of approximately $5\%$ compared to the results on the silicone dataset. 

The small CNN had a decrease of approximately $7\%$. Due to its simpler architecture, the small CNN model exhibited poorer generalization performance on the new dataset. This justifies our choice of using a state-of-the-art vision classifier. 

On the realistic dataset, the F1 scores when the models were fine-tuned on ground truth labels were lower than when fine-tuned on the MTurk labels. Analysis of the video revealed that the chicken skin would plastically deform during 
\begin{figurehere}
\begin{center}
\centerline{\includegraphics[width=\linewidth]{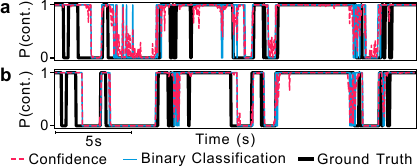}}
\caption{\newstuff{Predicted contact probabilities on one demonstration from the realistic dataset for EfficientNetB3 models  trained using (a) ground truth contact labels from force sensor measurements, and (b) human contact labels.}}
\label{figure:contact-time-realistic}
\end{center}
\vspace{-2em}
\end{figurehere}
manipulation. Thus, there were instances when the end effector would be grasping the chicken skin, but the forces as measured by the force sensor were low enough for a ``no-contact" classification. Under these conditions, the human labels were more accurate and less noisy than the ``ground truth" force sensor-based classifications. This observation explains the decrease in F1 scores on the realistic dataset compared to the silicone dataset for models trained on MTurk labels. In this scenario the false positive rate (as measured by precision in Table\,\ref{tab:contact_performance_silicone}) increased. The effect of this contact uncertainty in the force sensor labels can be seen in Fig.\,\ref{figure:contact-time-realistic}. Here, the model that was trained on force sensor contact labels (Fig.\,\ref{figure:contact-time-realistic}a) displayed more fluctuation in prediction confidence compared to the model that was trained on MTurk labels (Fig.\,\ref{figure:contact-time-realistic}b). The latter also showed more ``false positives" as defined by the ground truth, which was based on contact derived from force sensor data.

As indicated in Table \ref{tab: nn_rmse}, the position estimation methods retained similar performance levels as observed in the silicone dataset. Thus minimal fine-tuning was required to achieve good performance. This indicates that the proposed keypoint-based approach to position estimation exhibits data efficiency. \newstuff{A sample of the rescaled position estimate is shown in Fig.\,\ref{figure:nn-plot-realistic}.} 


\newstuff{The average estimated stiffness $k$ for the realistic dataset are listed in Table \ref{tab: stiffness}. The difference in mean stiffness between \textproc{C$_{\textproc{FS}}$--K$_{\textproc{FS}}$} and \textproc{C$_{\textproc{V}}$--K$_{\textproc{PSM}}$} were -32, -12, +39, -6\,$\text{Nm}^{-1}$ in the $X$, $Y$, $Z^+$, and $Z^-$ directions, respectively. Thus, the average error was 19\% across all directions with a maximum error of 41\% in the Z$^+$ direction. The low stiffness of the chicken skin in the Z$^+$ direction made stiffness estimates more sensitive to the noisy device dynamics. Thus, in its current form, the contact conditional force estimation methods have limited applicability to differentiation tasks involving very soft tissues.}

\begin{figure*}
\begin{center}
\centerline{\includegraphics[width=\linewidth]{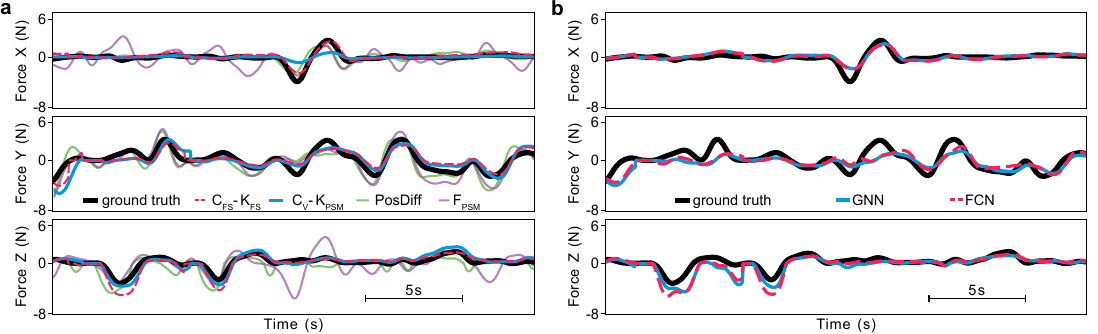}}
\caption{\newstuff{Example force predictions for force estimation approaches that require (a) robot state, and (b) no robot state information, for one demonstration from the realistic dataset.}}
\label{figure:M4_L_V1_1}
\end{center}
\vspace{-2em}
\end{figure*}

Consistent with earlier findings on the silicone dataset, Table \ref{tab: rmse_silicone} demonstrates that \textproc{C$_{\textproc{V}}$--K$_{\textproc{PSM}}$} yielded a lower average NRMSE compared to both joint torque-based force readings and \textproc{PosDiff}. The marginal increase in error observed between \textproc{C$_{\textproc{FS}}$--K$_{\textproc{FS}}$} and \textproc{C$_{\textproc{V}}$--K$_{\textproc{FS}}$} was significantly lower than the discrepancy seen between \textproc{C$_{\textproc{V}}$--K$_{\textproc{FS}}$} and 
\begin{figurehere}
\begin{center}
\centerline{\includegraphics[width=\linewidth]{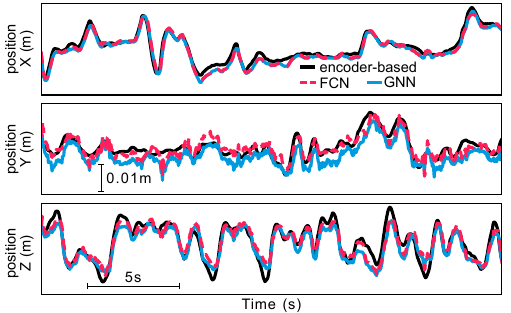}}
\caption{\newstuff{Scaled position predictions using the graph neural network (GNN) and fully connected network (FCN) compared to the joint encoder-based position estimated of the end effector on the realistic dataset.}}
\label{figure:nn-plot-realistic}
\end{center}
\vspace{-2em}
\end{figurehere}
\textproc{C$_{\textproc{V}}$--K$_{\textproc{PSM}}$}. Similar to the silicone dataset, this pattern indicates the large error contribution of K$_{\textproc{PSM}}$. The high Z force error is explained by the high error in the fitted stiffness constants in that direction. The high Y force error is due to occurrences of poor stiffness fits at the individual demonstration level. This was in part due to the plastic deformation of the chicken skin identified earlier in this section. The contact would be detected, but zero force would be exerted on the chicken skin, leading to erroneous stiffness measurements. The poor stiffness fits were partially masked within the aggregate computation of the mean stiffness. One possible approach to reducing the impact of this issue is to use a prior known tissue stiffness. This stiffness can then be conditionally updated during or after completion of the demonstration. Despite the relatively degraded stiffness estimates, \textproc{C$_{\textproc{V}}$--K$_{\textproc{PSM}}$} generally tracks force variation effectively, with the same trends as that of the silicone dataset as seen in Fig.\,\ref{figure:M4_L_V1_1}.


\subsubsection{Data Efficiency Experiments}

For the contact estimation, we considered the EfficientNet model pre-trained on the MTurk contact labels from the silicone dataset. We then fit models that were fine-tuned with increasing amounts of MTurk labels from the realistic dataset. The results presented in Fig.\,\ref{figure:EfficientNet_diff_size} showed that mean contact prediction accuracy began at a low of 85\% when the model was fine-tuned on only 50 additional examples. It gradually approached 90\% as the number of fine-tuning examples increased. The largest gains were seen between the range of 50 to 150 examples. This suggests that a very low amount of additional labeled fine-tuning data is required to approach peak performance on a new dataset.

For position estimation, we hypothesized that \newstuff{our model's abstract keypoint representation enables zero-shot transfer. Thus, we evaluated performance of the FCN and GNN models initially pre-trained exclusively on silicone}
\begin{figurehere}
\begin{center}
\centerline{\includegraphics[width=\linewidth]{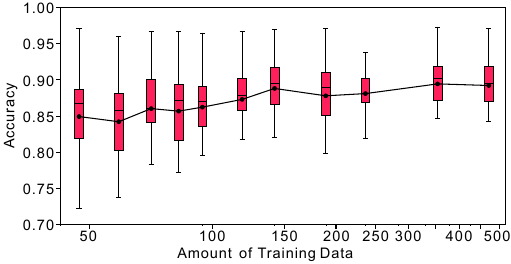}}
\caption{\newstuff{Box plot showing the accuracy of contact predictions using the EfficientNet-based visual contact detector trained on different numbers of human-labeled examples from the realistic dataset.  } The model was pre-trained on the silicone dataset. Connected dots represent the mean.\\}
\label{figure:EfficientNet_diff_size}
\end{center}
\vspace{-2em}
\end{figurehere}
\newstuff{data. They were subsequently fine-tuned with up to 2200 additional examples of end effector position data. } For all fine-tuning datasets, we re-state that the number of examples used to fine-tune DeepLabCut keypoint identification was only 90 images. The results shown in Fig.\,\ref{figure:pos-efficiency}a indicate that without additional fine-tuning, the FCN performed better than the GNN. Fine-tuning the FCN on only 200 examples from the realistic dataset results in performance that approaches the performance of models that used the full realistic training set. While the GNN showed less generalizability to a novel dataset, it can also be fine-tuned on a small amount of data, requiring approximately 300 additional examples to approach peak accuracy. Thus, our results suggest that the deeper and less constrained FCN has stronger representational flexibility. Therefore, it showcased better suitability for zero-shot transfer and fine-tuning. 

On the other hand, the GNN is better suited to novel deployments from scratch. This makes useful in clinical contexts where very little training data exists. Fig.\,\ref{figure:pos-efficiency}b shows the data-efficiency of the FCN and GNN models when trained from scratch (without pre-training on the silicone dataset). Here, the GNN had quicker convergence to peak accuracy than the FCN. Our results are thus consistent with other studies of fine-tuning for transfer learning to clinical-like data \cite{itzkovichGeneralizationDeepLearning2022}.

\subsection{Future Work}

In this work, we did not consider the influence of trocar forces on the resultant joint torque estimates of the robot. These forces can be significant and thus affect the accuracy of fitting local stiffness models based on torque estimates. Compared to end effector force sensing, trocar force sensing is more feasible to implement, given that the requirements for miniaturization and biocompatibility are less strict \cite{Fontanelli2020trocar}. Future work will study the feasibility of using trocar-based force sensing to augment our force estimation approach or learn a compensation model.

The use of an existing dataset did not allow us to fit dynamic models of the dVRK. Such models require custom calibration routines to be run on a specific robot \cite{fontanelli2017modelling,Wang2019dynamic}. The use of these dynamic models would likely improve the accuracy of stiffness estimates that we derive from the robot state information. 

Our reliance on normalized position estimates learned through a GNN or FCN, though suitable for \newstuff{clinical } data, provides an opportunity for future improvement. Performance of our position estimation can be enhanced by fitting precision camera models for a stereo endoscope, vision-based estimation algorithms \cite{haoVisionBasedSurgicalTool2018, SuPer}, or further developing learning-based 3D reconstruction methods like Neural Radiance Fields \cite{wangNeuralRenderingStereo2022}. Future research will also look into deeper GNN architectures. These would aim to leverage both the geometric graph structure that leads to data-efficient learning, and the deeper layers that were featured in the FCN. 

One area of improvement and further research for the contact-conditional approach is to account for slip. This can
\begin{figurehere}
\begin{center}
\centerline{\includegraphics[width=\linewidth]{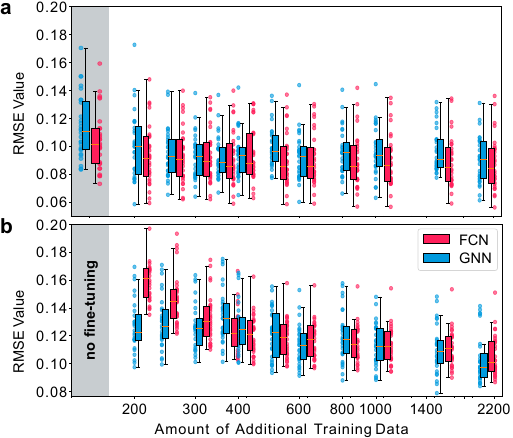}}
\caption{Box plots showing RMSE of normalized position estimates for models trained on different amounts of human-labeled examples from the realistic dataset with (a) pre-training on silicone dataset, and (b) without pre-training. Colored dots indicate individual demonstrations.}
\label{figure:pos-efficiency}
\end{center}
\vspace{-2em}
\end{figurehere}
be obtained via visual estimation or through in-built slip detection capabilities \cite{burkhardSlipSensingIntelligent2018}. Additionally, non-linear stiffness models can be used to improve the quality of force estimation. These can provide critical force information when tissues are stretched to high displacements that might induce tearing. 

Lastly, while we motivate the application of our work by considering video-based surgical skill evaluation, we did not conduct user studies to verify the construct validity of the force measures that we derive from our approach. Thus, future work will include user studies that compute various automated performance measures of tissue handling skill \cite{trejosDevelopmentForcebasedMetrics2014}, based on contact-conditional vision-based force estimates. \newstuff{Testing of these estimates for both direct and sensory substitution force feedback will also be conducted to evaluate potential benefits for real-time telesurgical manipulation.}


\section{Conclusion}

In this work we present an approach to hybrid model- and learning-based approach to visual force estimation. Unlike traditional supervised learning methods, the approach does not rely on external sensor measurements for model training and parameter fitting. The contact detection and keypoint-based labeling leverages human crowd-sourcing, which we verify to have comparable accuracy to sensor-based labels. \newstuff{The accuracy of our methods makes them highly applicable in tissue handling skill evaluations and for providing haptic feedback via sensory substitution.

We demonstrate that our approach has the added advantage of being quickly adaptable and scalable to novel scenarios. The developed learning-based normalized position estimator exhibits zero-shot transfer capability to new scenarios. Furthermore, its performance can be further improved via fine-tuning on end effector position measurements. The learning-based position estimator consequently enables contact-conditional force estimation for video-only surgical data streams. This key feature makes our methods highly suitable for clinical settings, where data is often limited.}

\nonumsection{Acknowledgments}
\noindent This work made use of the High Performance Computing Resource in the Core Facility for Advanced Research Computing at Case Western Reserve University. The authors would like to thank Owen Queeglay at Stanford University for helping to collect the realistic dataset. 



\bibliographystyle{ieeetr}
\bibliography{biblio}

{\bf Shuyuan Yang} received his B.Eng. degree from the
Taiyuan University of Technology, in 2020. He is currently a Master's Candidate at Case Western University where his thesis work is on studying machine learning approaches to vision-based robot pose position estimation.
\\ \\
{\bf My Le} is a Bachelor of Science student in the Department of Electrical, Computer, and Systems Engineering.
\\ \\
{\bf Kyle Golobish} received his B.S. in Mechanical Engineering from Case Western Reserve University. He currently works as a mechanical designer at Neuronoff Inc.
\\ \\
{\bf Juan Beaver} is a Bachelor of Science student in the Department of Electrical, Computer, and Systems Engineering.
\\ \\ 
{\bf Zonghe Chua} received the M.S. and Ph.D. degrees from Stanford University, Stanford, CA, in 2020 and 2022, respectively, all in mechanical engineering. He is currently an Assistant Professor of Electrical Engineering at Case Western Reserve University, Cleveland, OH, where he directs the Enhanced Robotic Interfaces and Experience Lab. He is a member of the IEEE, the Robotics and Automation Society, the Technical Committee on Haptics, and the Technical Committee on Telerobotics. His research interests include teleoperation, robot-assisted surgery, haptic feedback, and machine learning approaches for augmented human-robot interfaces.

\end{multicols}
\end{document}